\newcommand{\method}{EDAS\xspace}

\documentclass[nohyperref]{article}

\usepackage{microtype}
\usepackage{graphicx}
\usepackage{subfigure}
\usepackage{booktabs} 
\usepackage[font=small,labelfont=bf,
   justification=justified,
   format=plain]{caption}

\usepackage{hyperref}
\usepackage[dvipsnames]{xcolor}

\usepackage[ruled,vlined,linesnumbered,algo2e]{algorithm2e}
\DontPrintSemicolon
\SetAlgoLined

\usepackage[accepted]{icml2022}

\usepackage{amsmath}
\usepackage{amssymb}
\usepackage{mathtools}
\usepackage{amsthm}
\usepackage[overload]{empheq}
\usepackage[capitalize,noabbrev]{cleveref}

\theoremstyle{plain}

\theoremstyle{definition}

\theoremstyle{remark}

\usepackage[textsize=tiny]{todonotes}

\usepackage{etoolbox}
\patchcmd{\thebibliography}
  {\clubpenalty 4000\widowpenalty 4000}
  {\clubpenalties 1 10000 \widowpenalties 1 10000 }
  {}{}

\icmltitlerunning{Submission and Formatting Instructions for ICML 2022}

\begin{document}

\twocolumn[
\icmltitle{Equivariant Data Augmentation for Generalization\\in Offline Reinforcement Learning}




\begin{icmlauthorlist}
\icmlauthor{Cristina Pinneri*}{yyy}
\icmlauthor{Sarah Bechtle}{comp}
\icmlauthor{Markus Wulfmeier}{comp}
\icmlauthor{Arunkumar Byravan}{comp}
\icmlauthor{Jingwei Zhang}{comp}
\icmlauthor{William F. Whitney}{comp}
\icmlauthor{Martin Riedmiller}{comp}
\end{icmlauthorlist}

\icmlaffiliation{yyy}{Max Planck ETH Center for Learning Systems, T\"ubingen (Germany), Z\"urich (Switzerland)}
\icmlaffiliation{comp}{Google DeepMind, London, UK}

\icmlcorrespondingauthor{Cristina Pinneri}{cristina.pinneri@tuebingen.mpg.de}

\icmlkeywords{Machine Learning, ICML}

\vskip 0.3in
]



\printAffiliationsAndNotice{*Work done during internship at Google DeepMind.}


\begin{abstract}
We present a novel approach to address the challenge of generalization in offline reinforcement learning (RL), where the agent learns from a fixed dataset without any additional interaction with the environment. Specifically, we aim to improve the agent's ability to generalize to out-of-distribution goals. To achieve this, we propose to learn a dynamics model and check if it is equivariant with respect to a fixed type of transformation, namely translations in the state space. We then use an entropy regularizer to increase the equivariant set and augment the dataset with the resulting transformed samples. Finally, we learn a new policy offline based on the augmented dataset, with an off-the-shelf offline RL algorithm. Our experimental results demonstrate that our approach can greatly improve the test performance of the policy on the considered environments.
\end{abstract}
\section{Introduction}

Offline reinforcement learning (RL), also known as batch RL, is a paradigm that proposes to learn from an external source of data, without requiring any additional interaction between the agent and the environment, effectively decoupling the collection process from the inference stage in reinforcement learning \cite{riedmiller2022collect, siegel2020keep}, and making it possible to learn in real-word scenarios, where interactions with the environment are oftentimes costly or dangerous \cite{levine2020offline, prudencio2023survey}. Offline RL is also a valid alternative to learning dynamics models from logged data, especially when the underlying dynamics are complex, uncertain, or noisy. In such cases, direct model-based approaches might struggle to accurately represent the environment, and offline RL provides a robust way to learn effective policies.
Furthermore, offline RL is more versatile and flexible than its online variants since the training data can come from different distributions - e.g. expert demonstrations, near-optimal policies, random actions. 
In practice, offline RL comes with several challenges. One significant issue is the difficulty to obtain enough data from the optimal state-action distribution, problem known as \emph{distribution mismatch}. For model-free approaches based on the estimation of the Q function, this usually results in overestimation errors that lead to poor generalization performance \cite{kumar2020cql, wang2020crr}.
\begin{figure}[t]
    \centering
    \includegraphics[width=0.45\textwidth]{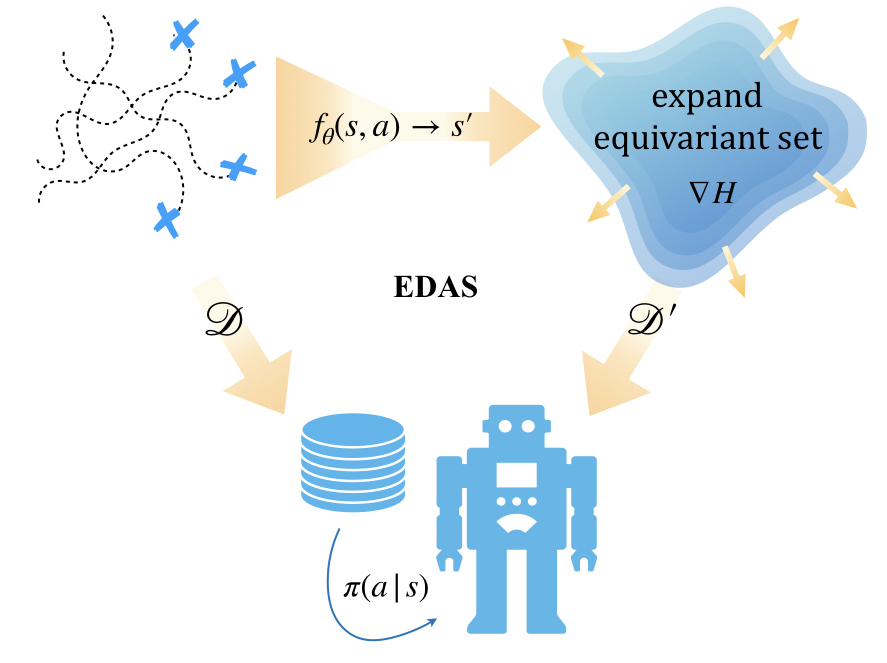}
    \vspace{-0.5em}
    \caption{We introduce EDAS: an algorithm for Equivariant Data Augmentation from State-inputs for generalization in offline RL. In step 1 we train a dynamical model from logged trajectories. In step 2 we use the trained model to generated a synthetic dataset of points which are equivariant to it. In the last step, the augmented dataset is used to train a generic offline RL algorithm.} 
    \vspace{-0.5em}
    \label{fig:process}
\end{figure}

In this paper, we explore the intersection of model-free and model-based offline RL, leveraging the strengths of both paradigms to address out-of-distribution (OOD) generalization. We propose a novel framework that combines data augmentation techniques with dynamical model learning. 
Data augmentation has proven to be a powerful technique for improving the performance of machine learning models, particularly in image classification tasks. Common forms of data augmentation for images include translations, reflections, rotations, and random cropping. These transformations are applied because they preserve the label of the image for virtually all types of images. However, when dealing with robotic systems, we cannot automatically apply these transformations, as we do not know which invariances are present in our dataset or if they only hold in certain parts of the data space. If in computer vision we want to find augmentations which are \emph{label-preserving}, in control tasks we want them to be \emph{physics-preserving}. 
In the context of decision-making, such as robots interacting within an environment to solve a task, these data augmentations could also be hard-coded, e.g., if we already know the system has some symmetries. Alternatively, they can be learned. In this paper, we explore the latter direction and propose a framework that, given a specified group transformation $\mathcal{G}$ (in our case, translations $T$), automatically finds the equivariant set and augments the dataset accordingly. 

We present \textbf{\method{}}, Equivariant Data Augmentation from State-inputs for offline RL: A plug-and-play data augmentation technique for improving the generalization performance of goal-directed policies in offline RL. \method{} exploits possible equivariances of the dynamics model, and incorporates a regularizer based on entropy to learn the parameters of the augmentation transformation. In this way, \method{} expands the given dataset in a principled way, improving the generalization performance of the policy.
We test \method{} through a proof of concept on two low-dimensional goal-conditioned control tasks, and an experiment with noisy dynamics. Our proposed approach serves as a stepping stone towards a rigorous understanding of the underlying mechanisms. We acknowledge the need of conducting additional research to further validate and refine the methodology, and extend the framework to higher-dimensional systems and other domains, such as pixel inputs. 

\section{Related Work}
Our work intersects with several areas of research, including model-free and model-based offline RL, data augmentation techniques, and equivariant learning. We provide an overview of some of these methods and position our work in this context.

Invariances and equivariances are incorporated in deep learning frameworks in different ways: at the architectural level \cite{cohen2016group, vanderpol2020homomorphic}, when learning data representations \cite{anselmi2019symmetry, zhang2021learning, salter2021attentionprivileged}, and in data augmentation \cite{dao2019kernel, tobin2017domain}. Data augmentation has proven to be an efficient and simple way for reducing overfitting and improving out-of-distribution generalization in classification tasks from images \cite{krizhevsky2017imagenet, he2016deep, ratner2017learning, antoniou2018data, cubuk2019autoaugment, shorten2019survey}.
Less work has been done on data augmentation for reinforcement learning tasks from images and observations \cite{laskin2020reinforcement,yarats2021image,sinha2022s4rl, ye2020rotation}. In particular, both RAD \cite{laskin2020reinforcement} and S4RL \cite{sinha2022s4rl} propose transformation distributions for state-inputs with fixed hyperparameters. Moreover, the magnitude of the transformation for state-inputs is very small ($3 \cdot 10^{- 4}$ in S4RL) which improves robustness through random perturbations rather than substantial data augmentation for generalization in OOD states.
Data augmentation has also been used in goal-conditioned RL tasks, with methods such as Hindsight Experience Replay (HER) \cite{andrychowicz2017hindsight}. Even though it was originally proposed to address the problem of sparse rewards, it can be seen as a form of data-augmentation for state inputs, as it generates synthetic goals from failed trajectories. 
In this work, we hinge on the generalization potential of model-based learning \cite{SuttonBarto1998,Sun2018ModelbasedRI,dong20d, Young23benefitsmbrl} and propose a method to learn the augmentation's parameters in order to achieve maximal generalization, leveraging the equivariance property of the learned dynamical model. 
\begin{figure*}[h]
  \centering
  \begin{minipage}[t]{0.3\textwidth}
    \includegraphics[width=\linewidth]{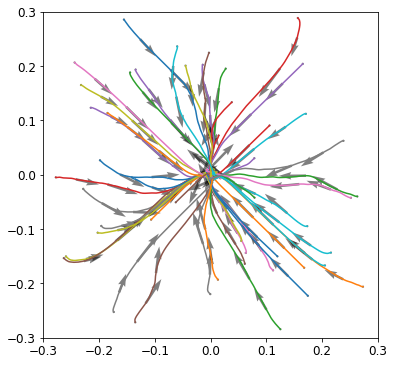}\vspace{-1.5em}
    \makeatletter
    \renewcommand{\fnum@figure}{(a)}
    \makeatother
    \captionof*{figure}{Original training dataset. The trajectories all end in $(0,0)$ with zero velocity. Notice the arrow direction.}
    \label{fig:pm_traj_orig}
  \end{minipage}\hfill
  \begin{minipage}[t]{0.3\textwidth}
    \includegraphics[width=\linewidth]{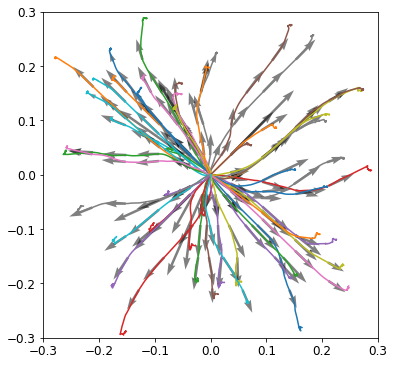}\vspace{-1.5em}
    \makeatletter
    \renewcommand{\fnum@figure}{(b)}
    \makeatother
    \captionof*{figure}{Trajectories are subtracted a constant offset determined by their starting point. Now all the trajectories start at the origin.}
    \label{fig:pm_traj_minus_start}
  \end{minipage}\hfill
  \begin{minipage}[t]{0.3\textwidth}
    \includegraphics[width=\linewidth]{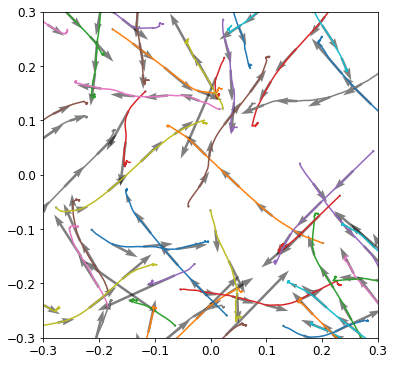}\vspace{-1.5em}
    \makeatletter
    \renewcommand{\fnum@figure}{(c)}
    \makeatother
    \captionof*{figure}{Every trajectory gets shifted by a random offset sampled from $U[l,h]$, where $l$ and $h$ are the bounds of the arena.}
    \label{fig:pm_traj_minus_rnd}
  \end{minipage}\\[1em]
    \begin{minipage}[t]{0.3\textwidth}
    \includegraphics[width=\linewidth]{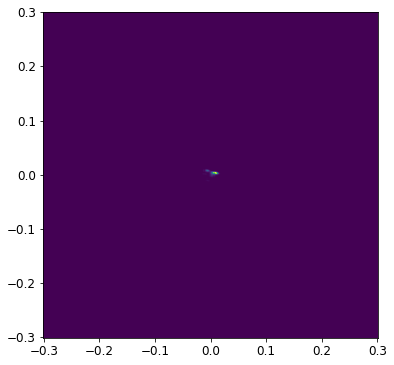}\vspace{-1.5em}
    \makeatletter
    \renewcommand{\fnum@figure}{(d)}
    \makeatother
    \captionof*{figure}{Goal distribution $\rho_g$ for the original training dataset. The only point with zero velocity (terminal state) is $(x,y)=(0,0)$. Goal relabeling with HER does not influence $\rho_g$.}
    \label{fig:pm_orig_goal}
  \end{minipage}\hfill
  \begin{minipage}[t]{0.3\textwidth}
    \includegraphics[width=\linewidth]{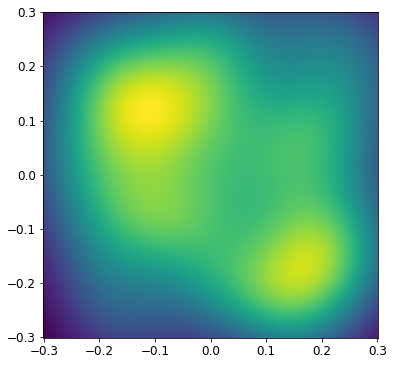}\vspace{-1.5em}
    \makeatletter
    \renewcommand{\fnum@figure}{(e)}
    \makeatother
    \captionof*{figure}{Since the coordinate system is now relative to the starting state of each trajectory, the goal distribution is not centered around zero anymore.}
    \label{fig:pm_orig_minus_start_goal}
  \end{minipage}\hfill
  \begin{minipage}[t]{0.3\textwidth}
    \includegraphics[width=\linewidth]{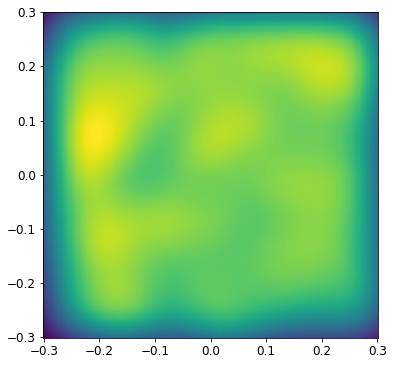}\vspace{-1.5em}
    \makeatletter
    \renewcommand{\fnum@figure}{(f)}
    \makeatother
    \captionof*{figure}{The final states are spread all over the state space and their distribution is closer to a uniform distribution.}
    \label{fig:pm_orig_minus_rnd_goal}
  \end{minipage}\\
    \caption{Data augmentation process and its effects on trajectories (upper row) and goal distribution (lower row) in the Point Mass environment: (a-c) demonstrate the translation and transformation steps applied to trajectories, while (d-f) showcase the evolution of the goal distribution $\rho_g$ throughout the process. We could apply random translations to the Point Mass environment because we know that the system has translational invariance. \method{} learns this transformation automatically, increasing the bounds of the uniform distribution $U_\phi$ to have maximum entropy.}  \label{fig:pm_dataaug_process}
\end{figure*}
\section{EDAS}
\subsection{Intuitive Example}
Before discussing the components of our method, we present an intuitive explanation of the procedure in the Point Mass environment (Fig.~\ref{fig:pm_dataaug_process}).
We consider a training dataset of 100 trajectories that start from random points inside the arena and end up at its center in $(0,0)$, as shown in Fig.~\ref{fig:pm_dataaug_process}~(a). Notably, the training set contains only one terminal state, the center.
However, to enable the policy to effectively generalize to different terminal states, we need to modify the final goal distribution. We achieve this through invariant transformations that respect the physical laws governing the system. For the Point Mass environment, shifting the coordinate system's origin to the trajectory's starting point yields a new dataset (Fig.~\ref{fig:pm_dataaug_process}~(b)). This alteration redistributes terminal states, no longer peaked in (0,0), as shown in Fig.~\ref{fig:pm_dataaug_process}~(e).

Intuitively, maximizing the entropy of the terminal state distribution enhances generalization. We attain this by uniformly sampling the origin of the coordinate system (Fig.~\ref{fig:pm_dataaug_process} (c)). With \method{}, we introduce a method to autonomously learn these transformations, ensuring that the generating distribution maximizes entropy while adhering to the system's properties.

\subsection{Methods}
We operate in the usual Markov Decision Process framework defined by the tuple $\mathcal{M}=<\mathcal{S},\mathcal{A},\mathcal{G}, \mathcal{R}, \mathcal{P}, \rho_{s_0}>$. 
Equivariance refers to the property of a function commuting under the action of a group transformation \cite{cohen2016group}. The equivariance property can be forma\-lized as:
\begin{equation}
    f(g(x)) = g(f(x)) \quad \forall g \in \mathcal{G}, \forall x \in \mathcal{X}
\end{equation}
where $g \in \mathcal{G}$ is the group function, which can act on both domain and co-domain of the function $f$.

If we consider the group of translations in the state space, described by the group $\mathcal{T}$, the result of the translation is an addition operation:
\begin{align}\label{eq:translational_equivariance}
    f(t_u(x)) &= t_u(f(x)) \quad \forall t_u \in \mathcal{T}_u, \forall x \in \mathcal{X} \nonumber\\
    &\iff \nonumber\\
    f(x + u) &=f(x) + u
\end{align}
where $\mathcal{T}_u$ indicates translations by a fixed vector $u$. 
The purpose of this work is to learn the parameters of the distribution generating the fixed vectors $u$, so that the equivariance property in Eq.~\ref{eq:translational_equivariance} is respected. We call this distribution $U_\phi$, parameterized by $\phi$. 
The function $f$ is our learned dynamical model $f_\theta: \mathcal{S} \times \mathcal{A} \rightarrow \mathcal{S}$. This is used to increase the equivariant set with an entropy regularizer and use it to augment the original dataset by minimizing the following loss:
\begin{subequations}\label{eq:joint_loss_1stversion}
\begin{equation}
    \mathcal{L}_\phi = \mathcal{L}_{eq} - \lambda_e\mathcal{R}_{ent}+ \lambda_v \mathcal{R}_{val} \tag{\ref{eq:joint_loss_1stversion}} 
\end{equation}
\begin{align}[left={\empheqlbrace}]
    \mathcal{L}_{eq} &= \texttt{MSE}\left(f_\theta(u_\phi s,a) - u_\phi s'\right)\label{eq:equivariant_prediction_error} \\
    \mathcal{R}_{ent} &= H(U_\phi)\label{eq:entropy_regularizer} \\
    \mathcal{R}_{val} &= \texttt{ReLU}(\phi_l-\phi_h)
\end{align}
\end{subequations}
where the last regularization loss enforces the bounds of the uniform distribution to be valid (low $\phi_l$ $\leq$ high $\phi_h$).
The loss in Eq.~\ref{eq:joint_loss_1stversion} was originally introduced by \citet{benton2020augerino} for classification tasks but it can be extended to regression with dynamical models as well. In Section \ref{sec:noisy_dynamics} we propose a second loss in order to take into account imperfect models.

The pseudocode of the optimization process is illustrated in Alg.~\ref{algo:edas}. More specifically, given a dataset of transitions $\mathcal{D}=(s_i,a_i,r_i,s_i')$ with $i \in[1,N]$, we first learn a dynamical model $f_\theta(s,a)$ that maps a state-action pair to the next state; we then parameterize a uniform distribution $U$ with a parameter $\phi$ that comprises the upper and lower bounds of $U$, acting on each dimension of $\mathcal{S}$ separately. We aim to achieve equivariance of $f_\theta$ under the group transformation by minimizing the mean squared error between $f(u_\phi s,a)$ and $u_\phi s'$, where $u_\phi \in T_{U_\phi}$ denotes a random translation by a fixed vector $u$ sampled from $U_\phi$. Therefore, we encourage the expansion of the equivariant set with an entropy regularizer, i.e., we maximize the entropy of the distribution $U_\phi$, while making sure that the equivariance loss is minimized w.r.t.~$\phi$ (Eq.~\ref{eq:joint_loss_1stversion}). 
Finally, we obtain the augmented dataset $\mathcal{D}'= \cup_{i=1}^M T_{U_\phi}\mathcal{D}$ by applying $M$ translations to every point of $\mathcal{D}$, and use it to learn a new policy using offline RL algorithms. The number $M$ does not have to be bounded, as we can perform as many translations as we want until we have the desired goal coverage. 

The intuition behind this approach is that by learning an equivariant augmentation of the data, we can identify and exploit the invariances that are present in the system, which in turn can lead to better generalization performance. The interplay between the equivariance loss and the entropy regularizer ensures that the equivariant set will not expand if no invariance exists within the system.
The advantage of learning distribution bounds as in \citet{benton2020augerino}, rather than using fixed parameters as in \citet{laskin2020reinforcement}, \citet{sinha2022s4rl}, is that we can learn partial symmetries and represent systems that show invariances only in some parts of their domains.
\subsection{Extension to Noisy Dynamics}\label{sec:noisy_dynamics}
In practical settings we often deal with epistemic (e.g.~lack of data) or aleatoric sources of uncertainty (e.g.~noisy sensory readings).
This usually makes the generalization capabilities of the learned model less useful, because long term predictions are not reliable, and more sophisticated algorithms are required. 

We overcome this issue by considering the one-step equivariant prediction error like before (Eq.~\ref{eq:equivariant_prediction_error}) and, additionally, we compare it to the standard prediction error in this way:
\begin{equation}\label{eq:joint_loss_2ndversion}
    \mathcal{L} = \mathcal{R}_{ent} \cdot ( \mathcal{L}_{eq} - \mathcal{L}_{dyn} - \lambda_e) + \lambda_v\mathcal{R}_{val} 
\end{equation}
where $\mathcal{L}_{dyn} = \texttt{MSE}\left(f_\theta( s,a) - s'\right)$.
We limit ourselves to the 1-step prediction error and, in particular, how it compares to the accuracy of the model on the untransformed data ($\mathcal{L}_{eq} - \mathcal{L}_{dyn}$). This allows us to exploit the generalization power of the model without relying on inaccurate long-term predictions. If we only relied on the absolute error rather than its difference w.r.t.~the original value, we would risk increasing the entropy in areas subject to noise.
Multiplying the relative error to the entropy regularizer introduces more conservative updates of the equivariant set. The entropy is minimized if $( \mathcal{L}_{eq} - \mathcal{L}_{dyn} - \lambda_e)>0$, namely if we observe that equivariant predictions are less accurate than the untransformed predictions by a margin greater than $\lambda_e$. On the other side, when the losses are comparable, $( \mathcal{L}_{eq} - \mathcal{L}_{dyn} - \lambda_e)<0$, and the entropy can be safely maximized. 
\newlength\myindent 
\setlength\myindent{1em} 
\newcommand\bindent{%
  \begingroup 
  \setlength{\itemindent}{\myindent} 
  \addtolength{\algorithmicindent}{\myindent} 
}
\newcommand\eindent{\endgroup} 

\begin{algorithm}[htb]

\renewcommand{\baselinestretch}{2.0}

\SetAlgoLined
\caption{Equivariant Data Augmentation for Offline RL}
\textbf{Require} Dataset of transitions $\mathcal{D} = {(s, a, r, s')_i}$ with $i \in [1,N]$, transformation group $\mathcal{G}$ (e.g., translations), entropy regularization weight $\lambda$, batch size $\mathcal{B}$, number of training iterations $K$

\vspace{0.5em}
\SetKwFunction{learnmodel}{TrainDynamicsModel}
\SetKwProg{firstlossprocedure}{Step 1}{:}{}
\firstlossprocedure{\learnmodel}{
\vspace{0.2em}

$ \mathcal{L}_{dyn}=\frac{1}{|\mathcal{D}|}\sum_{(s, a, s') \in \mathcal{D}} ||f_\theta(s, a) - s'||^2$ 
}

\vspace{0.5em}
\SetKwFunction{expandset}{ExpandEquivariantSet}
\SetKwProg{secondlossprocedure}{Step 2}{:}{}
\secondlossprocedure{\expandset}{
\vspace{0.2em}

Initialize uniform distribution parameters $\phi$

\For{$k = 1, \ldots, K$}{

Sample a minibatch of transitions ${(s, a, s')}$ from $\mathcal{D}$

Sample a minibatch of offsets ${u_\phi}$ from $U_\phi$

Compute equivariance loss:\\
$\mathcal{L}_{\text{eq}} = \frac{1}{|\mathcal{B}|}\sum_{(s, a, s', u_\phi) \in \mathcal{B}} ||f_\theta(u_\phi( s), a) -u_\phi( s')||^2$

Compute entropy regularizer:\\
$\mathcal{R}_{\text{ent}} = H(U_\phi)$

Compute bound validity regularizer:\\
$\mathcal{R}_{\text{val}} = \texttt{ReLU}(\phi_l - \phi_h)$

Update $\phi$ using gradient descent:

$\phi \leftarrow \phi - \eta \nabla_\phi (\mathcal{L}_{\text{eq}} - \lambda_e \mathcal{R}_{\text{ent}} + \lambda_v \mathcal{R}_{val})$
}
Generate augmented dataset $\mathcal{D}' = \bigcup_{i=1}^M T_{U_\phi}\mathcal{D}$
}

\vspace{0.5em}
\SetKwFunction{offlineRL}{OfflineRL}
\SetKwProg{thirdlossprocedure}{Step 3}{:}{}
\thirdlossprocedure{\offlineRL}{
\vspace{0.2em}
Use any offline RL algorithm with the original dataset $\mathcal{D}$ and the 
augmented dataset $\mathcal{D}'$
}
\vspace{1em}
\KwResult{Learned policy $\pi$}
\label{algo:edas}
\end{algorithm}
\section{Results}
In this section, we present the results of our experiments to evaluate \method{}. We considered two tasks in the DMControl Suite \cite{tunyasuvunakool2020dmcontrol}: the Point Mass and the Planar Reacher (``\texttt{hard}''), illustrated in Fig.~\ref{fig:traintest} (a-b). Even though these environments can be easily solved, the learned policies struggle to generalize to unseen goals. 
Our aim is to learn the appropriate transformation bounds, so that we obtain equivalent test and train performance.

\subsection{Learned Bounds}\label{sec:bounds}

\begin{figure}[h]
    \centering
        \begin{minipage}[t]{0.22\textwidth}
    \makeatletter
    \renewcommand{\fnum@figure}{(a)}
    \makeatother
    \centering
    \includegraphics[width=\linewidth]{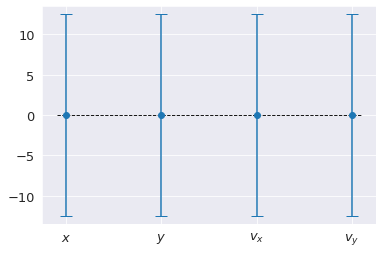}\vspace{-0.5em}
    \captionof*{figure}{Bounds explode if we only maximize the entropy ($\mathcal{R}_{ent}$).}
    \label{fig:pmbounds_onlyentropy}
    \end{minipage}\hfill
    \begin{minipage}[t]{0.23\textwidth}
    \makeatletter
    \renewcommand{\fnum@figure}{(b)}
    \makeatother
    \centering
    \includegraphics[width=\linewidth]{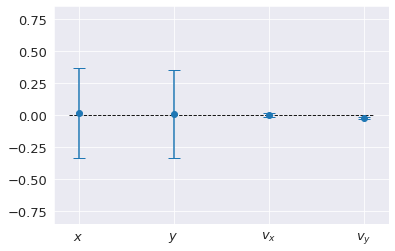}\vspace{-0.5em}
    \captionof*{figure}{Bounds with equivariant prediction error and $\mathcal{R}_{ent}$.}
    \label{fig:pmbounds_plusequivariantloss}
    \end{minipage}
    \caption{Learned transformation bounds (upper and lower bounds of uniform distribution) for the Point Mass task.}\label{fig:pmbounds}
\end{figure}

\paragraph{\textsc{Point Mass}} The state space consists of Cartesian positions and velocities ($x,y,v_x,v_y$), and the dynamics model is linear. If we apply a translation to the positions, the velocities will remain unchanged:
$$
\begin{cases}
   x'(t) =& u_\phi x(t) = \boldsymbol{x(t)+ u} \\
    \dot{x}'(t) =& \frac{\delta}{\delta t} u_\phi x(t)= \frac{\delta}{\delta t}  (x(t) + u) = \boldsymbol{\dot{x}(t)}\\
\end{cases}
$$
Therefore, in order to have perfect equivariance, we cannot apply a transformation on the velocities. 
This is confirmed by our results in Fig.~\ref{fig:pmbounds} (b) where the learned translation gets applied only on the positions, and the learned upper and lower bound correspond to the arena limits, while the bounds for the velocity get compressed to zero.

\begin{figure}[h]
    \centering
    \includegraphics[width=0.6\linewidth]{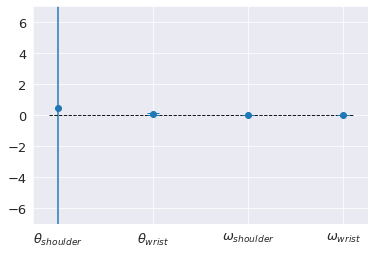}\vspace{-0.5em}
    \caption{Learned transformation bounds (upper and lower bounds of uniform distribution) for the Planar Reacher task (hard).}\label{fig:reacherbounds}
\end{figure}

\paragraph{\textsc{Planar Reacher}} In this environment, a two-link robotic arm aims to reach a target in a two-dimensional plane. The state space consists of joint angles -- $\theta_{shoulder}$ and $\theta_{wrist}$ -- and their angular velocities, and the dynamics model is nonlinear. If we add an offset to the shoulder angle (translation), this results in a rotation of the arm around the origin. This transformation is equivariant w.r.t.~the dynamics, which holds because the Planar Reacher is not affected by gravity, otherwise this translational symmetry would be broken and other affine transformations should be considered (e.g.~reflectional). The learned bounds for this task, as seen in Fig.~\ref{fig:reacherbounds} only apply arbitrary translations to the shoulder angle.
\subsection{Offline RL}\label{sec:offlinerl}
\begin{figure}[h]
    \centering
        \begin{minipage}[t]{0.22\textwidth}
    \makeatletter
    \renewcommand{\fnum@figure}{(a)}
    \makeatother
    \centering
    \includegraphics[width=0.8\linewidth]{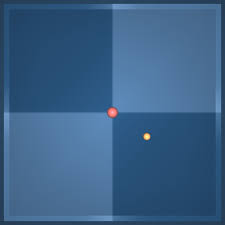}\vspace{-0.5em}
    \captionof*{figure}{Point Mass}
    \label{fig:point_mass}
    \end{minipage}\hfill
    \begin{minipage}[t]{0.22\textwidth}
    \makeatletter
    \renewcommand{\fnum@figure}{(b)}
    \makeatother
    \centering
    \includegraphics[width=0.8\linewidth]{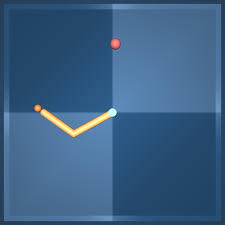}\vspace{-0.5em}
    \captionof*{figure}{Planar Reacher}
    \label{fig:reacher}
    \end{minipage}
    \bigskip

    \begin{minipage}[t]{0.22\textwidth}
    \makeatletter
    \renewcommand{\fnum@figure}{(c)}
    \makeatother
    \centering
    \includegraphics[width=\linewidth]{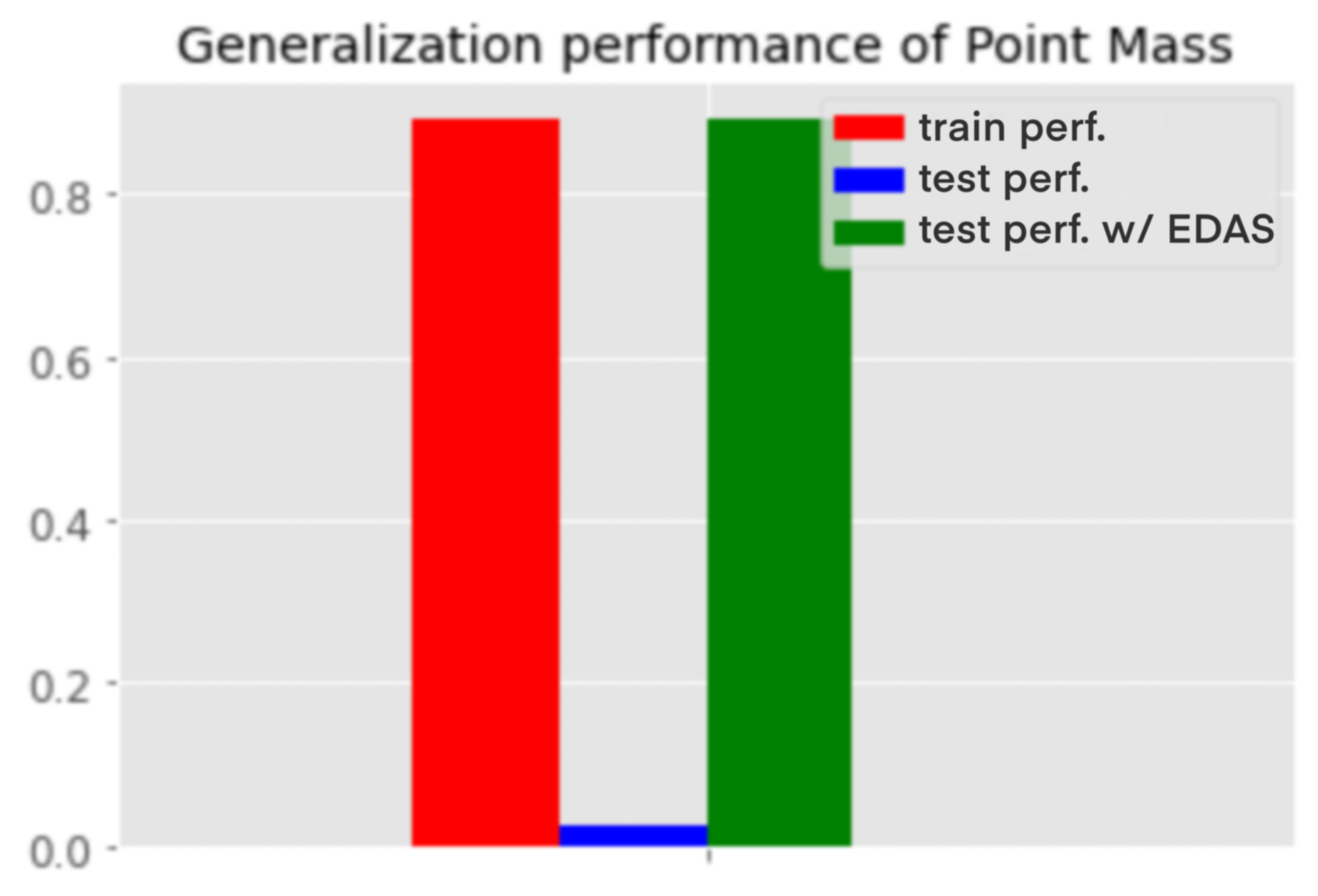}
    \captionof*{figure}{The training dataset for the Point Mass task consisted of 100 trajectories all terminating in $(0,0)$. 
    When tested on different goals, the performance dropped to zero.}
    \label{fig:point_mass_improvement}
    \end{minipage}\hfill
    \begin{minipage}[t]{0.22\textwidth}
    \makeatletter
    \renewcommand{\fnum@figure}{(d)}
    \makeatother
    \centering
    \includegraphics[width=\linewidth]{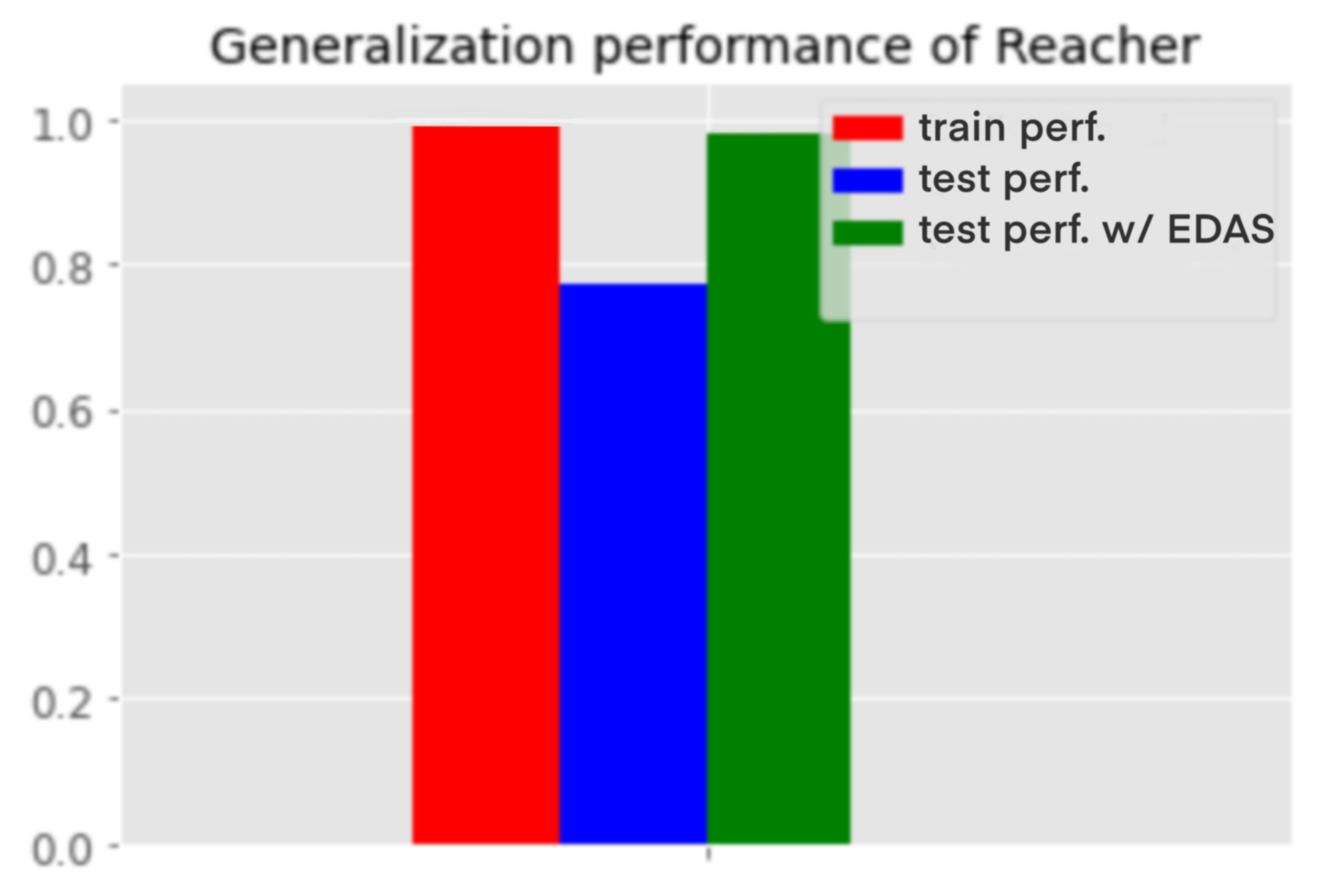}
    \captionof*{figure}{The training goals for the Planar Reacher were sampled uniformly in the state space. When tested on different goals, the performance degraded.}
    \label{fig:reacher_improvement}
\end{minipage}
\caption{Average return, normalized from 0 to 1. {\color{Red} Red}: performance on goals seen during training. {\color{Blue} Blue}: performance on test goals. {\color{OliveGreen} Green}: performance on test goals after training on the augmented dataset generated by \method{}. In this case, there is no substantial difference between train and test performance.}\label{fig:traintest}
\end{figure}

After learning the bounds, we constructed the augmented dataset $\mathcal{D}'$ by repeatedly sampling offsets from $U_\phi$ and applying them to the original dataset. This step could be iterated until a sufficient goal coverage is reached. Both datasets, $\mathcal{D} \cup \mathcal{D}'$, were used to train a goal-conditioned policy with a generic offline reinforcement learning algorithm (CRR by  \citet{wang2020crr}, with a temperature equal to 1 in our case). The learned policy demonstrated improved test performance when compared to results on the original dataset without augmentation. In particular, in Fig.~\ref{fig:traintest} we observe that the test and train performances are essentially the same, demonstrating the effectiveness of our approach for generalization to unseen situations.


\subsection{Imperfect Models}\label{sec:imperfect_models}
\begin{figure}[h]
    \centering
        \begin{minipage}[t]{0.22\textwidth}
    \makeatletter
    \renewcommand{\fnum@figure}{(a)}
    \makeatother
    \centering
    \includegraphics[width=0.7\linewidth]{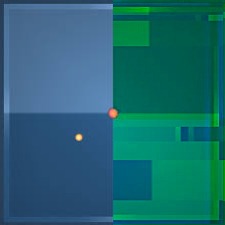}\vspace{-0.5em}
    \captionof*{figure}{Noisy Point Mass: uniform noise is applied to all the data points with $x>0$.}
    \label{fig:noisypointmass}
    \end{minipage}\hfill
    \begin{minipage}[t]{0.24\textwidth}
    \makeatletter
    \renewcommand{\fnum@figure}{(b)}
    \makeatother
    \centering
    \includegraphics[width=\linewidth]{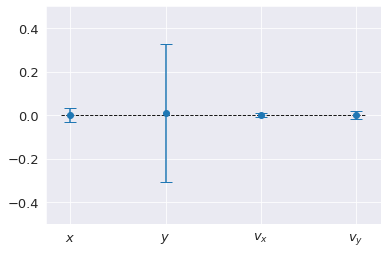}\vspace{-0.5em}
    \captionof*{figure}{Translations along the $x$-axis are now highly discouraged.}
    \label{fig:noisypmbounds}
    \end{minipage}
    \caption{Learned transformation bounds (upper and lower bounds of uniform distribution) for the noisy Point Mass task.}\label{fig:imperfectmodel}
\end{figure}

We conducted an experiment on the Point Mass environment under partially noisy dynamics. In this case, we apply some noise to all the training data that have a positive $x$-coordinate (Fig.~\ref{fig:imperfectmodel} (a)). The added noise disrupts the model's ability to make consistent and accurate predictions. This scenario also simulates situations where an unknown source of disturbance affects only specific regions of the state space. 
We train \method{} with the modified equation that takes into account error differences (Eq.~\ref{eq:joint_loss_2ndversion}).
The results in Fig.~\ref{fig:imperfectmodel} (b) show that translations along the $x$-axis are now penalized because, whenever we add a random vector $u$ that shifts the Point Mass to the right side, we get a bigger prediction error ($\mathcal{L}_{eq}$) than the one on the unshifted data ($\mathcal{L}_{dyn}$). Our formulation with 1-step relative errors (Eq.~\ref{eq:joint_loss_2ndversion}) allows for a more conservative behaviour that takes into account the dynamics model's accuracy.

\section{Conclusion and Future Work}
In this work we proposed \method{}, a framework for data augmentation in offline RL that leverages potential equivariances of the dynamical model to increase the generalization performance of the learned policy. Our approach is shown to be effective in improving the robustness of the learned policy to changes in the task and environment, reaching nearly equal test and train performance. While our study represents an exploratory step in understanding the application of equivariant representations in offline RL, our results suggest that it could be a promising direction for future research.

Possible directions for future work include considering different generating distributions for the fixed vectors $u$. Currently, the set of vectors is sampled from a hypercube defined by the multivariate uniform distribution $U$. If the system presents non-trivial invariances, more expressive distributions could be necessary. 
Additionally, it is important to investigate the impact of other affine transformations -- such as reflection, rotation, and scaling \cite{vanderpol2020homomorphic,wang2022mathrmsoequivariant} -- and regularization strategies on the performance of our method. Moreover, we will also investigate the extension of our approach to higher-dimensional control tasks. This is in principle possible for any system that exhibits an invariance w.r.t.~one or more joint variables \emph{independently}. 

Another potential avenue for future research is to extend this work to the online setting, and explore how the equivariance prediction error can be used to guide exploration during the model learning process, allowing us to test whether the system is invariant to certain types of transformations or not. This could provide valuable insights into the underlying dynamics of the system and help improve the accuracy of the learned model.
Future work could also focus on using stochastic models and integrating uncertainty into the equivariance loss, particularly in cases where the loss value is low when evaluated on the model output but not on its uncertainty. Additionally, analyzing a joint framework where the model and policy are trained end-to-end could lead to a tighter integration between model-free and model-based equivariant RL.



In summary, our proposed framework \method{} demonstrates the value of leveraging the equivariant prediction error to improve policy generalization performance. We believe that the integration of model equivariance into reinforcement learning algorithms could open up many research directions, and contribute to the development of more efficient, robust, and adaptable RL systems.

\bibliography{main}
\bibliographystyle{icml2022}



\end{document}